\pdfoutput=1
\documentclass[11pt,a4paper]{article}
\usepackage[hyperref]{two-column-style}
\usepackage{times}
\usepackage{latexsym}
\usepackage{color}
\usepackage{amsmath,amssymb}
\usepackage[utf8]{inputenc}

\usepackage{url}

\usepackage{blindtext}
\usepackage{booktabs}

\usepackage{todonotes}
\usepackage{microtype}
\usepackage{rotating}
\usepackage[normalem]{ulem}
\usepackage{enumitem}

\newcommand{\xldrop}[1] {\scriptsize{(#1)}}
\newcommand{\xlgain}[1] {\scriptsize{(-#1)}}

\aclfinalcopy

\title{
Concatenated Power Mean Word Embeddings \\as Universal Cross-Lingual Sentence Representations
}

\author{
Andreas R\"uckl\'e\textsuperscript{$\dagger$}, \hspace{0.4mm}
Steffen Eger\textsuperscript{$\dagger$}, \hspace{0.4mm}
Maxime Peyrard\textsuperscript{$\dagger\ddagger$}, \hspace{0.4mm}
Iryna Gurevych\textsuperscript{$\dagger\ddagger$}
\\[.3em]
	\textsuperscript{$\dagger$}Ubiquitous Knowledge Processing Lab (UKP)\\
    	\textsuperscript{$\ddagger$}Research Training Group AIPHES\\
	Department of Computer Science, Technische Universit\"{a}t Darmstadt\\
	\textsuperscript{$\dagger$} {\tt www.ukp.tu-darmstadt.de} \\ 
    \textsuperscript{$\ddagger$} {\tt www.aiphes.tu-darmstadt.de}\\
}

\date{}

\begin{document}
\maketitle
\begin{abstract}
Average word embeddings are a common baseline for more sophisticated sentence embedding techniques.
However, they typically fall short of the performances 
of more complex models such as InferSent.
Here, we generalize the concept of average word embeddings to \emph{power mean word embeddings}.
We show that the concatenation of different types of power mean word embeddings considerably closes the gap to state-of-the-art methods
\emph{monolingually} and substantially outperforms these more complex techniques \emph{cross-lingually}.  
In addition, our proposed method outperforms different recently proposed baselines such as SIF and Sent2Vec by a solid margin, thus constituting a much harder-to-beat monolingual baseline.
Our data and code are publicly available.\footnote{\url{https://github.com/UKPLab/arxiv2018-xling-sentence-embeddings}}
\end{abstract}

\section{Introduction}\label{sec:introduction}
Sentence embeddings are dense vectors that summarize different properties of a sentence (e.g.\ its meaning),
thereby extending the very popular concept of word embeddings \cite{Mikolov:2013,Pennington2014} to the sentence level.

\emph{Universal sentence embeddings} have recently gained considerable attention due to their wide range of possible applications in downstream tasks.
In contrast to task-specific representations, such as the ones trained specifically for tasks like textual entailment or sentiment,
such sentence embeddings
are 
trained in a task-agnostic manner on large datasets.
As a consequence, 
they often perform better
when little labeled data is available 
\cite{Subramanian2018}.

To a certain degree, the history
of sentence embeddings
parallels that of word embeddings, but on a faster scale:
early word embeddings models were complex and often took months to train 
\cite{Bengio:2003,Collobert:2008,Turian:2010}
before \newcite{Mikolov:2013} presented a much simpler
method that could train substantially faster and therefore on much more data, leading to significantly better results. Likewise, sentence embeddings originated from the rather resource-intensive `Skip-thought' encoder-decoder model of \newcite{Kiros:2015}, before successively less demanding models 
\cite{Hill2016,Kenter:2016,Arora2017} 
were proposed that are much faster at train and/or test time.

The most popular state-of-the-art approach is the so-called InferSent model \cite{Conneau2017}, which learns
sentence embeddings with a rather simple architecture in single day (on a GPU), but on very high quality data, namely, Natural Language Inference data \cite{Bowman:2015}. Following previous work (e.g.\ \citealt{Kiros:2015}), 
InferSent has also set the standards in measuring the usefulness of sentence embeddings
by requiring 
the embeddings 
to be
``universal'' in the sense that they must yield stable and high-performing results on a wide variety of so-called ``transfer tasks''.  

We follow both of these trends and posit that sentence embeddings should be simple, on the one hand, and universal, on the other. Importantly, we extend universality to the cross-lingual case: universal sentence embeddings should perform well across multiple tasks \text{and} across natural languages. 

The arguably simplest 
sentence embedding technique is to average individual word embeddings. This so-called 
mean word embedding is the starting point of our extensions. 

First, we observe that average word embeddings have partly been treated unfairly in previous work such as \newcite{Conneau2017} because the newly proposed methods yield sentence embeddings of rather large size (e.g., $d=4096$) while they have been compared to 
much smaller average word embeddings (e.g., $d=300$). 
Increasing the size of individual---and thus average---word embeddings is likely to improve the  
quality of average word embeddings, but with an inherent limitation: 
there is (practically) only a finite number of words
in natural languages, so that the additional dimensions will not be used to store additional information, beyond a certain threshold. To remedy, (i) we instead propose to concatenate diverse word embeddings that store \emph{different} kinds of information, such as syntactic, semantic or sentiment information; concatenation is a  
simple but effective technique 
in different setups \cite{Zhang:2016}.

Secondly, and more importantly, (ii) we posit that
`mean' has been defined too narrowly by the corresponding NLP community. Standard mean 
word embeddings
stack the word vectors of a sentence in a matrix
and compute per-dimension \emph{arithmetic} means on this matrix. We perceive this mean as a summary of all the entries in a dimension. 
In this work, we instead focus on 
\emph{power means} \cite{Hardy:1952}
which naturally generalize
the arithmetic mean.

Finally,
(iii) we combine concatenation of word embeddings with different power means and show that our sentence embeddings satisfy our requirement of universality: 
they substantially outperform different other strong baselines across a number of tasks monolingually, and substantially outperform 
other approaches cross-lingually.

\section{Related Work}\label{sec:related}

\paragraph{Monolingual word embeddings} 
are typically learned to predict context words in fixed windows \cite{Mikolov:2013,Pennington2014}. 
Extensions predict contexts given by dependency trees \cite{Levy:2014} or combinations of windows and dependency context \cite{Komninos:2016}, leading to more syntactically oriented word embeddings. Fasttext \cite{Bojanowski2017} represents words as the sum of their n-gram representations trained with a skip-gram model.  
Attract-repel \cite{Mrksic2017} uses synonymy
and antonymy constraints from lexical
resources to fine tune word embeddings with linguistic
information.
\newcite{Vulic2017} morph-fit word embeddings using language-specific rules so that
derivational antonyms (``expensive'' vs.\ ``inexpensive'') move far away in vector space.

\paragraph{Cross-lingual word embeddings} originate from the idea that not only monolingually but also cross-lingually similar words should be close in vector space. 
Common practice is to learn a mapping between two monolingual word embedding spaces \cite{faruqui-dyer:2014:EACL,artetxe2016learning}.
Other approaches predict mono- and bilingual context using word alignment information as an extension to the standard skip-gram model \cite{Luong:2015} or inject cross-lingual synonymy and antonymy constraints similar as in the monolingual setting \cite{Mrksic2017}. As with monolingual embeddings, there exists a veritable zoo of different approaches, but they have been reported to nonetheless often perform similarly in applications \cite{Upadhyay2016}.

In this work, we train one of the simplest approaches: BIVCD \cite{Vulic2015}. This creates bilingual word embeddings from aligned bilingual documents by concatenating parallel document pairs and shuffling the words in them before running a standard word embedding technique.

\paragraph{Monolingual sentence embeddings} usually built on top of existing word embeddings, and different approaches focus on computing sentence embeddings by composition of word embeddings. \citet{Wieting:TACL} learned paraphrastic sentence embeddings by fine-tuning skip-gram word vectors while using additive composition to obtain representations for short phrases. SIF \cite{Arora2017} computes sentence embeddings by taking weighted averages 
of word embeddings and then modifying them via SVD.
Sent2vec \cite{Pagliardini2017} learns n-gram embeddings and averages them.
Siamese-CBOW \cite{Kenter:2016} trains word embeddings that, when averaged, should yield good representations of sentences.
However, even non-optimized average word embeddings can encode valuable information about the sentence, such as its length and word content \cite{Adi2017}.

Other approaches consider sentences as additional tokens whose embeddings are learned jointly with words \cite{Le2014}, use auto-encoders \cite{Hill2016}, or mimick the skip-gram model 
\cite{Mikolov:2013} 
by predicting surrounding sentences \cite{Kiros:2015}.

Recently, InferSent \cite{Conneau2017} achieved state-of-the-art results across a wide range of different transfer tasks. Their model uses bidirectional LSTMs and was trained on the SNLI \cite{Bowman:2015} and MultiNLI \cite{Williams2017} corpora. 
This is novel in 
that previous work, which  
likewise used LSTMs to learn sentence embeddings but trained
on other tasks (i.e.\ identifying paraphrase pairs), usually did not achieve significant improvements compared to simple word averaging models \cite{Wieting16}.

\paragraph{Cross-lingual sentence embeddings} have received comparatively less attention. 
\citet{Hermann2014} learn cross-lingual word embeddings and infer document-level representations with simple composition of unigrams or bigrams, finding that added word embeddings perform on par with the more complex bigram model. Several authors proposed to extend ParagraphVec \cite{Le2014} to the cross-lingual case: \newcite{Pham2015} 
add a bilingual constraint to learn cross-lingual representations using aligned sentences; \citet{mogadala-rettinger:2016} add a general cross-lingual regularization term to ParagraphVec;
\citet{Zhou2016} train task-specific representations for sentiment analysis based on ParagraphVec by minimizing the distance between paragraph embeddings of translations.
Finally, \citet{Chandar2013} train a cross-lingual auto-encoder to learn representations that allow reconstructing sentences and documents in different languages, and
\citet{Schwenk2017} use representations learned by an NMT model for translation retrieval.

To our best knowledge, all of these cross-lingual works evaluate on 
few
individual datasets, 
and none focuses on
\emph{universal} cross-lingual sentence embeddings that perform well across a wide range of different tasks.

\section{Concatenated Power Mean Embeddings}

\paragraph{Power means}\label{sec:model}
Our core idea is generalizing average word embeddings,
which summarize a sequence of embeddings $\mathbf{w}_1,...,\mathbf{w}_n\in\mathbb{R}^d$ by component-wise arithmetic averages:
\begin{align*}
  \forall i=1,\ldots,d:\:\:\frac{w_{1i}+\cdots+w_{ni}}{n}
\end{align*}
This operation summarizes the `time-series' $(w_{1i},\ldots,w_{ni})$ of variable length $n$ by their arithmetic mean. Of course, then, we might also compute other statistics on these time-series such as standard deviation, skewness (and further moments), Fourier transformations, etc., in order to 
capture different information from the sequence.

For simplicity and to focus on only one type of extension, we consider in this work so-called \emph{power means} \cite{Hardy:1952}, defined as:
\begin{align*}
  \left(\frac{x_1^p+\cdots+x_n^p}{n}\right)^{1/p};\quad p\in\mathbb{R}\cup\{\pm\infty\}
\end{align*}
for a sequence of numbers $(x_1,\ldots,x_n)$. 
This generalized form retrieves many well-known means such as 
the arithmetic mean ($p=1$), the geometric mean ($p=0$), and the harmonic mean ($p=-1$). In the extreme cases, when $p=\pm \infty$, the power mean specializes to the minimum ($p=-\infty$) and maximum ($p=+\infty$) of the sequence.

\paragraph{Concatenation}
For %a sequence of 
vectors $\mathbf{w}_1,\ldots,\mathbf{w}_n$, concisely written as a matrix $\mathbf{W}=[\mathbf{w}_1,\ldots,\mathbf{w}_n]\in\mathbb{R}^{n\times d}$, we let $H_p(\mathbf{W})$ stand for the vector in $\mathbb{R}^d$ whose $d$ components are the power means of the sequences $(w_{1i},\ldots,w_{ni})$, for all $i=1,\ldots,d$. 

Given a sentence $s=w_1\cdots w_n$ we first look up the embeddings $\mathbf{W}^{(i)}=[\mathbf{w}_1^{(i)},\ldots,\mathbf{w}_n^{(i)}]\in\mathbb{R}^{n\times d_i}$ of its words from some embedding space $\mathbb{E}^i$. To get 
%good
summary statistics of the sentence, we then compute $K$ power means of $s$ and concatenate them:
\begin{align*}
  \mathbf{s}^{(i)}=H_{p_1}(\mathbf{W}^{(i)})\oplus\cdots\oplus H_{p_K}(\mathbf{W}^{(i)})
\end{align*}
where $\oplus$ stands for concatenation and $p_1,\ldots,p_K$ are $K$ different power mean values. Our resulting sentence representation, denoted as $\mathbf{s}^{(i)}=\mathbf{s}^{(i)}(p_1,\ldots,p_k)$, lies in $\mathbb{R}^{d_i\cdot K}$.

To get further representational power from different word embeddings, we concatenate %these 
different power mean sentence representations $\mathbf{s}^{(i)}(p_1,\ldots,p_k)$ obtained from different embedding spaces $\mathbb{E}^i$:
\begin{align}\label{eq:main}
  \bigoplus_i\mathbf{s}^{(i)}
\end{align}
The dimensionality of this representation is $K\sum_i d_i$. When all embedding spaces have the same dimensionality $d$, this becomes $K\cdot L\cdot d$, where $L$ is the number of spaces considered.

\section{Monolingual Experiments}
\label{sec:mono-exp}

\subsection{Experimental Setup}

\paragraph{Tasks}
We replicate the setup of \newcite{Conneau2017} and evaluate on the six transfer tasks listed in their table 1.
Since their selection of tasks is slightly biased towards sentiment analysis, we add three further tasks:
\texttt{AM}, an argumentation mining task based on \newcite{Stab:2017} where sentences are classified into the categories major claim, claim, premise, and non-argumentative; \texttt{AC}, a further argumentation mining 
task with very few data points based on \newcite{Peldszus2016} in which the goal is to classify sentences as to whether they contain a claim or not;
and \texttt{CLS}, a task based on \newcite{Prettenhofer2010} to identify \emph{individual sentences} as being part of a positive or negative book review.\footnote{
The original CLS was built for \emph{document} classification.}

We summarize the different tasks in Table \ref{table:evaluation:tasks}.

\begin{table*}
\centering
\footnotesize
\begin{tabular}{llcccl}
\toprule
\textbf{Task} & \textbf{Type} & \textbf{Size} & \textbf{X-Ling} & \textbf{C} & \textbf{Example (X-Ling)} \\ 
\midrule
\texttt{AM} & Argumentation & 7k  & \hspace{1.6mm}\textsf{HT}$^\dagger$ & 4 & Viele der technologischen Fortschritte helfen der Umwelt sehr.  \textit{(claim)} 
 \\
\texttt{AC} & Argumentation & 450 & \hspace{1.6mm}\textsf{HT}$^\dagger$ & 2 & Too many promises have not been kept. \textit{(none)} \\
\texttt{CLS} & Product-reviews & 6k & \textsf{HT} & 2 & En tout cas on ne s'ennuie pas à la lecture de cet ouvrage! \textit{(pos)} \\
\midrule 
\texttt{MR} & Sentiment & 11k & \textsf{MT} & 2 & Dunkel und verstörend, aber auch überraschend witzig. \textit{(pos)} \\
\texttt{CR} & Product-reviews & 4k & \textsf{MT} & 2 & This camera has a major design flaw. \textit{(neg)} \\
\texttt{SUBJ} & Subjectivity & 10k & \textsf{MT} & 2 & On leur raconte l'histoire de la chambre des secrets. \textit{(obj)} \\
\texttt{MPQA} & Opinion-polarity & 11k & \textsf{MT} & 2 & sind eifrig \textit{(pos)} $|$ nicht zu unterstützen \textit{(neg)} \\
\texttt{TREC} & Question-types & 6k & \textsf{MT} & 6 & What's the Olympic Motto? \textit{(desc---question asking for description)} \\
\texttt{SST} & Sentiment & 70k & \textsf{MT} & 2 & Holm... incarne le personnage avec un charisme regal [...] \textit{(pos)} \\
\bottomrule
\end{tabular}
\caption{Evaluation tasks with examples from our transfer languages. The first three tasks include human-generated cross-lingual data (\textsf{HT}), the last 6 tasks contain machine translated sentences (\textsf{MT}). C is the number of classes. \\$^\dagger$ indicates that a dataset contains machine translations for French.}
\label{table:evaluation:tasks}
\end{table*}

\paragraph{Word embeddings}
We use four diverse, potentially complementary types of word embeddings
as basis for our sentence representation techniques:
\texttt{GloVe} embeddings (GV) \cite{Pennington2014} trained on Common Crawl;
Word2Vec \cite{Mikolov2013} trained on \texttt{GoogleNews} (GN); \texttt{Attract-Repel} (AR) \cite{Mrksic2017} 
and \texttt{MorphSpecialized} (MS) \cite{Vulic2017}.

We use pre-trained word embeddings except for Attract-Repel where we use the retrofitting code 
from
the authors
to tune the embeddings of \newcite{Komninos:2016}.

\paragraph{Evaluated approaches}
For each type of word embedding, we evaluate the standard average 
($p=1$) as sentence embedding as well as different power mean concatenations.
We also evaluate concatenations of embeddings $\mathbf{s}^{(i)}(1,\pm \infty)$, where $i$ ranges over the 
word embeddings mentioned above.\footnote{Monolingually, we limit our experiments to the three named power means to not exceed the dimensionality of InferSent.}
We motivate this choice of power means later in our analysis. %in \S\ref{sec:analysis}.

We compare 
against the following four 
approaches: \texttt{SIF} \cite{Arora2017}, 
applied to GloVe vectors; average \texttt{Siamese-CBOW} embeddings \cite{Kenter:2016} based on the Toronto Book Corpus;
\texttt{Sent2Vec} \cite{Pagliardini2017}, 
and \texttt{InferSent}. 

While SIF ($d=300$), average Siamese-CBOW ($d=300$), and Sent2Vec ($d=700$) embeddings are relatively low-dimensional, InferSent embeddings are high-dimensional ($d = 4096$). In all our experiments the maximum dimensionality of our concatenated power mean sentence embeddings does not exceed $d=4\cdot 3\cdot 300=3600$.

\paragraph{Evaluation procedure}

We train a 
logistic regression classifier on top of sentence embeddings 
\emph{for our added tasks} 
with 
random subsample validation 
(50 runs)
to mitigate the effects of different random initializations.
We use 
SGD with Adam and 
tune the learning rate on the validation set.
In contrast, for a direct comparison against previously published results, we use 
SentEval \cite{Conneau2017} to evaluate MR, CR, SUBJ, MPQA, TREC, and SST. For most tasks, this approach likewise uses logistic regression with %10-fold 
cross-validation. 

We report macro F1 performance for AM, AC, and CLS to account for imbalanced classes, and accuracy for all tasks evaluated using SentEval.

\subsection{Results}
\begin{table*}[t]
\centering
\footnotesize
\begin{tabular}{l|c|cccccccccc}
\toprule
\textbf{Model} & $\Sigma$ & \textbf{AM} & \textbf{AC} & \textbf{CLS} & \textbf{MR} & \textbf{CR} & \textbf{SUBJ} & \textbf{MPQA} & \textbf{SST} & \textbf{TREC} \\ 
\midrule 
\multicolumn{11}{l}{\textbf{Arithmetic mean}} \\
\midrule
GloVe (GV) & 77.2 & 50.0 & 70.3 & 76.6 & 77.1 & 78.3 & 91.3 & 87.9 & 80.2 & 83.4 \\
GoogleNews (GN) & 76.1 & 50.6 & 69.4 & 75.2 & 76.3 & 74.6 & 89.7 & 88.2 & 79.9 & 81.0  \\
Morph Specialized (MS) & 73.5 & 47.1 & 64.6 & 74.1 & 73.0 & 73.1 & 86.9 & 88.8 & 78.3 & 76.0 \\
Attract-Repel (AR) & 74.1 & 50.3 & 63.8 & 75.3 & 73.7 & 72.4 & 88.0 & 89.1 & 78.3 & 76.0 \\
%GV $\oplus$ GN & 78.3 & 52.8 & 71.0 & 76.8 & 77.9 & 78.6 & 91.6 & 88.6 & 81.5 & 86.2  \\ 
%GV $\oplus$ GN $\oplus$ MS & 78.7 & 53.5 & 70.9 & 77.0 & 77.9 & 79.6 & \textbf{91.9} & 88.9 & 81.6 & 86.6 \\
GV $\oplus$ GN $\oplus$ MS $\oplus$ AR & \textbf{79.1} & \textbf{53.9} & \textbf{71.1} & \textbf{77.2} & \textbf{78.2} & \textbf{79.8} & 91.8 & \textbf{89.1} & \textbf{82.8} & \textbf{87.6} \\
\midrule
\multicolumn{11}{l}{\textbf{power mean}  \scriptsize{\lbrack p-values\rbrack}} \\
\midrule
GV \scriptsize{$\lbrack -\infty, 1, \infty\rbrack$} & 77.9 & 54.4 & 69.5 & 76.4 & 76.9 & 78.6 & 92.1 & 87.4 & 80.3 & 85.6  \\
GN \scriptsize{$\lbrack -\infty, 1, \infty\rbrack$} & 77.9 & 55.6 & 71.4 & 75.8 & 76.4 & 78.0 & 90.4 & 88.4 & 80.0 & 85.2  \\
MS \scriptsize{$\lbrack -\infty, 1, \infty\rbrack$} & 75.8 & 52.1 & 66.6 & 73.9 & 73.1 & 75.8 & 89.7 & 87.1 & 79.1 & 84.8 \\
AR \scriptsize{$\lbrack -\infty, 1, \infty\rbrack$} & 77.6 & 55.6 & 68.2 & 75.1 & 74.7 & 77.5 & 89.5 & 88.2 & 80.3 & 89.6 \\
GV $\oplus$ GN $\oplus$ MS $\oplus$ AR \scriptsize{$\lbrack -\infty, 1, \infty\rbrack$} & 80.1 & 58.4 & 71.5 & 77.0 & 78.4 & 80.4 & \underline{\textbf{93.1}} & 88.9 & 83.0 & 90.6 \\
~~~ $\rightarrow$ with z-norm$^\dagger$ & \textbf{81.1} & \textbf{60.5} & \underline{\textbf{75.5}} & \textbf{77.3} & \textbf{78.9} & \textbf{80.8} & 93.0 & \textbf{89.5} & \textbf{83.6} & \underline{\textbf{91.0}} \\
\midrule
\multicolumn{11}{l}{\textbf{Baselines}} \\
\midrule
GloVe + SIF & 76.1 & 45.6 & 72.2 & 75.4 & 77.3 & 78.6 & 90.5 & 87.0 & 80.7 & 78.0 \\
Siamese-CBOW & 60.7 & 42.6 & 45.1 & 66.4 & 61.8 & 63.8 & 75.8 & 71.7 & 61.9 & 56.8 \\
Sent2Vec & 78.0 & 52.4 & \textbf{72.7} & 75.9 & 76.3 & 80.3 & 91.1 & 86.6 & 77.7 & \textbf{88.8} \\
InferSent & \underline{\textbf{81.7}} & \underline{\textbf{60.9}} & 72.4 & \underline{\textbf{78.0}} & \underline{\textbf{81.2}} & \underline{\textbf{86.7}} & \textbf{92.6} & \underline{\textbf{90.6}} & \underline{\textbf{85.0}} & 88.2 \\
%~~~ $\rightarrow$ with normalization$^\dagger$ & 81.6 & 61.1 & 73.0 & 78.0 & 81.3 & 85.6 & 92.5 & 90.7 & 84.1 & 87.8 \\

\bottomrule
\end{tabular}
\caption{Monolingual results. Brackets show the different power means that were applied to all individual embeddings. $^\dagger$ we normalized the embeddings of our full model with the z-norm as proposed by \newcite{LeCun:1998}.
}
\label{table:results:monolingual}
\end{table*}

Table \ref{table:results:monolingual} 
compares models
across all 9 transfer tasks.
The results show that we can substantially improve sentence embeddings when concatenating multiple word embedding types. 
All four embedding types concatenated achieve 2pp improvement over 
the best individual embeddings (GV). 
Incorporating further power means also substantially improve performances.
GV improves by 0.6pp on average, GN by 1.9pp, MS by 2.1pp and AR by 3.7pp when concatenating $p=\pm\infty$ to the standard value $p=1$ 
(dimensionality increases
to 900). 
The combination of concatenation of embedding types and power means gives an average improvement of 3pp over the individually best embedding type.

However, there is one caveat with concatenated power mean embeddings: both the concatenated embeddings as well as the different power means live in their own ``coordinate system'', i.e., may have different ranges. 
Thus, we subtract the column-wise mean of the embedding matrix as well as divide by the standard deviation, which is the z-norm operation as proposed in \cite{LeCun:1998}. This indeed improves the results by 1.0pp. 
We thereby reduce the gap to InferSent from 4.6pp to 0.6pp (or 85\%), while having a lower dimensionality (3600 vs 4096).
For InferSent, this normalization decreased scores by 0.1pp on average.

We consistently outperform the lower-dimensional SIF, Siamese-CBOW, and Sent2Vec embeddings. 
We conjecture that these methods underperform as universal sentence embeddings\footnote{SIF and others were originally only evaluated in textual similarity tasks.} because they each discard important information.
For instance, SIF assigns low weight to common words such as discourse markers, while Siamese-CBOW similarly tends to assign low vector norm to function words \cite{Kenter:2016}. However, depending on the task, function words may have crucial signaling value.
For instance, in AM, words like 
``thus'' often indicate argumentativeness.

While the representations learned by Siamese-CBOW and SIF are indeed lower-dimensional than both our own representations as well as those of InferSent, we find it remarkable that they both perform below the (likewise low-dimensional) GV baseline on average. Sent2Vec (700d) outperforms GV, but performs below the concatenation of GV and GN (600d). This challenges their statuses as hard-to-beat baselines when evaluated on many different transfer tasks.

We further note that our concatenated power mean word embeddings outperform much more resource-intensive approaches such as Skip-thought 
in 4 out of 6 common tasks reported in \citet{Conneau2017} 
and 
the neural MT (en-fr) system reported there in 5 of 5 common tasks.

\begin{figure}

%\ffigbox{
  \centering
  \input{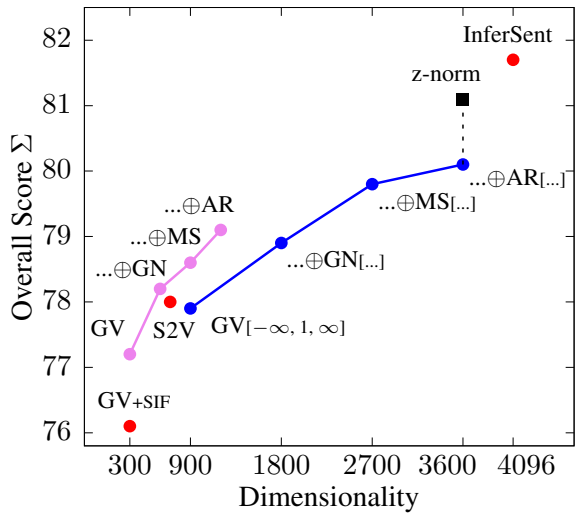}
%}{%
  \caption{The average monolingual performance for the different sentence embeddings in relation to their dimensionality. We visually group related embeddings (i.e.,  average and power mean embeddings). S2V is Sent2Vec.}
  \label{fig:experiments:plot}  
%}

\end{figure}

\paragraph{Dimensionality vs.\ Average Score} Our initial motivation %, we 
stated that a fair evaluation of sentence embeddings should compare embeddings of similar sizes.
Figure \ref{fig:experiments:plot} investigates 
the relationship of dimensionality and performance
based on our conducted experiments. We see that, indeed, larger embedding sizes lead to higher average performance scores; more precisely, there appears to be a sub-linear (logarithmic-like) growth in average performance as we increase embedding size through concatenation of diverse word embeddings. This holds for both the standard concatenation of average ($p=1$) embeddings and the $p$-mean concatenation with $p=1,\pm\infty$. 
Further, we observe that the concatenation of diverse average ($p=1$) word embeddings typically outperforms the $p$-mean summary of the same dimensionality ($p=1,\pm\infty$).
For example, concatenating arithmetic averages of GV, GN, and MS embeddings ($d=900$) outperforms the $p$-mean embeddings ($p=1,\pm\infty$) of GV.
%of the same size. 
Similarly, GV$\oplus$GN$ \oplus$MS$\oplus$AR ($d=1200$) outperforms the even higher-dimensional GV$\oplus$GN{\scriptsize$\lbrack 1,\pm\infty\rbrack$} score ($d=1800$). 
This suggests that there is a
trade-off for the considered $p$-mean concatenations: while they typically 
improve
performance, the increase is accompanied by an increase in embedding size, which makes alternatives (e.g., 
concatenation of arithmetic average embeddings)
competitive. However, when further concatenation of more embedding types is not possible (because no more are available)
or when re-training of a given embedding type with higher dimensionality is unfeasible (e.g., because the original resources are not available or because training times are prohibitive), concatenation of power mean embeddings offers a
strong
performance increase that is based on a better summary of 
the present
information.
Finally, we remark the very positive effect of z-normalization,
which
is
particularly beneficial in our situation of the concatenation of heterogeneous information; as stated, InferSent did not witness a corresponding performance increase. Our overall final result is very close to that of InferSent, while being lower dimensional and considerably cheaper at test time. At the same time, we do not rely on high quality inference data at train time, which is unavailable for most languages other than English. 

\section{Cross-Lingual Experiments}
\label{sec:xling-exp}
%In the monolingual evaluation we only measured one aspect of universal sentence embeddings, namely, 
%performance across transfer tasks.
%We now extend %our monolingual evaluation 
%to cross-lingual transfer.

\subsection{Experimental Setup}
\paragraph{Tasks}
We obtained German (de) and French (fr) translations of all 
sentences in our 9 transfer tasks.

Sentences in AC are already parallel (en, de), having been (semi-)professionally translated by humans from the original English. 
For AM, we 
use 
student translations from the original English into German \cite{Eger:2018}. 
CLS (en, de, fr) is also available bilingually. 
For the remaining datasets we created machine translated versions using Google Translate
for the directions en-de and en-fr.

\paragraph{Word embeddings}
Since our monolingual embeddings are not all available cross-lingually, we use alternatives:

\begin{itemize}[noitemsep,leftmargin=0.6cm]
\item We train en-de and en-fr \texttt{BIVCD} (BV) embeddings on aligned sentences from the Europarl \cite{Koehn:2005} and the UN corpus \cite{Ziemski:2016}, respectively, using word2vec; %\cite{Mikolov2013};
\item \texttt{Attract-Repel} (AR) \cite{Mrksic2017} provide pre-trained cross-lingual word embeddings for en-de and en-fr; 
\item 
Monolingual \texttt{Fasttext} (FT) word embeddings \cite{Bojanowski2017} of multiple languages 
trained on Wikipedia, which we map into shared vector space with a non-linear projection method similar to the ones proposed in \newcite{Wieting:TACL}, but with necessary modifications to account for the cross-lingual setting. (technical details 
are given in the appendix).
\end{itemize}
We also re-map the BV and AR embeddings using our technique. Even though BV performances were not affected by this projection, AR embeddings were greatly improved by it.  
All our cross-lingual word embeddings have $d=300$.

\paragraph{Evaluated approaches}

Similar to the monolingual case, we evaluate standard averages ($p=1$) for all embedding types, 
as well as different concatenations of word
embedding types and power means.
Since we have only three rather than four base embeddings here, 
we additionally report results for $p=3$. Again, we motivate our choice of $p$-means below. %in %\S\ref{sec:analysis}. 

We also evaluate bilingual SIF embeddings, i.e., SIF applied to bilingual word embeddings, 
%the 
CVM-add
of \citet{Hermann2014} with dimensionality $d=1000$ which we trained using sentences from Europarl and the UN corpus,\footnote{We observed that $d=1000$ performs slightly better than higher-dimensional CVM-add embeddings of $d=1500$ and much better than the standard configuration with $d=128$. This is in line with our assumption that single-type embeddings become better with higher dimension, but will not generate additional information beyond a certain threshold,  cf.~\S\ref{sec:introduction}.}
and 
three novel cross-lingual variants of InferSent:

(1) \texttt{InferSent MT}: We translated all 569K sentences in the SNLI corpus \cite{Bowman:2015} 
to German and French using Google Translate. 
To train, e.g., en-de InferSent, we consider all 4 possible language combinations over each sentence pair in the SNLI corpus.
Therefore, our new SNLI corpus is four times as large as the original.

(2) \texttt{InferSent TD}: We train the InferSent model on a different task where it has to differentiate between translations and unrelated sentences (translation detection), i.e., the model has two output classes but has otherwise the same architecture. To obtain translations, we use sentence translation pairs from Europarl (en-de) and the UN corpus (en-fr); unrelated sentences were randomly sampled from the respective corpora. We limited the number of training samples to the size of the SNLI corpus to keep the training time reasonable.\footnote{Also, adding 
more
data did not improve performances.} 

(3) \texttt{InferSent MT+TD}: This is a combination of the two previous approaches where we merge translation detection data with cross-lingual SNLI. The two label sets are combined, resulting in 5 different classes.

We trained all InferSent adaptations using the cross-lingual AR word embeddings.

We do not consider cross-lingual adaptations of ParagraphVec and NMT approaches 
here
because they already 
underperform 
simple 
word averaging models 
monolingually
\cite{Conneau2017}.

\paragraph{Evaluation procedure}
We replicate the 
monolingual evaluation procedure 
and train the classifiers on English sentence embeddings. 
However, we then measure the transfer performance on German and French sentences (en$\rightarrow$de, en$\rightarrow$fr). 

\subsection{Results}
%\paragraph{Results}
\begin{table*}[t]
\centering
\footnotesize
\begin{tabular}{l|c|cccccccccc}
\toprule
\textbf{Model} & $\Sigma$ & \textbf{AM} & \textbf{AC} & \textbf{CLS} & \textbf{MR} & \textbf{CR} & \textbf{SUBJ} & \textbf{MPQA} & \textbf{SST} & \textbf{TREC} \\ 
\midrule 
\multicolumn{11}{l}{\textbf{Arithmetic mean}} \\
\midrule
BIVCD (BV) & 67.3 & 40.5 & 67.6 & 66.3 & 64.4 & 71.7 & 81.1 & 81.6 & 65.7 & 67.0 \\
 & \scriptsize{(3.7)} & \scriptsize{(5.6)} & \scriptsize{(3.1)} & \scriptsize{(4.0)} &  \scriptsize{(1.9)} & \scriptsize{(0.6)} & \scriptsize{(3.5)} & \scriptsize{(3.1)} & \scriptsize{(3.8)} & \scriptsize{(7.7)}  \\
Attract-Repel (AR) & 69.2 & 38.6 & 68.8 & 68.9 & 68.2 & 73.9 & 82.8 & 84.4 & 72.5 & 64.5 \\
& \scriptsize{(3.6)} & \scriptsize{(4.8)} & \scriptsize{(0.8)} & \scriptsize{(4.3)} &  \scriptsize{(3.4)} & \scriptsize{(2.1)} & \scriptsize{(3.0)} & \scriptsize{(1.8)} & \scriptsize{(3.4)} & \scriptsize{(9.2)} \\
FastText (FT) & 68.3 & 38.4 & 63.4 & 70.0 & 69.1 & 73.1 & 85.1 & 81.5 & 69.3 & 65.1 \\
& \scriptsize{(5.6)} & \scriptsize{(8.5)} & \scriptsize{(2.9)} & \scriptsize{(4.1)} &  \scriptsize{(4.1)} & \scriptsize{(2.5)} & \scriptsize{(3.6)} & \scriptsize{(4.5)} & \scriptsize{(8.6)} & \scriptsize{(11.7)} \\
%BV $\oplus$ AR & 71.1 & \textbf{41.5} & \textbf{68.8} & 70.1 & 68.9 & 75.9 & 84.0 & \textbf{84.7} & \textbf{74.1} & \textbf{71.9} \\
%& \scriptsize{(4.2)} & \scriptsize{(8.6)} & \scriptsize{(3.1)} & \scriptsize{(3.8)} &  \scriptsize{(3.4)} & \scriptsize{(0.7)} & \scriptsize{(3.4)} & \scriptsize{(2.8)} & \scriptsize{(4.1)} & \scriptsize{(7.5)} \\
BV $\oplus$ AR $\oplus$ FT & \textbf{71.2} & 40.0 & 67.7 & \underline{\textbf{71.6}} & \textbf{70.3} & \underline{\textbf{76.8}} & \textbf{86.2} & 84.7 & 73.3 & 70.5 \\
& \scriptsize{(5.9)} & \scriptsize{(11.8)} & \scriptsize{(3.3)} & \scriptsize{(3.6)} &  \scriptsize{(5.1)} & \scriptsize{(1.4)} & \scriptsize{(3.8)} & \scriptsize{(3.3)} & \scriptsize{(6.8)} & \scriptsize{(13.9)} \\
\midrule
\multicolumn{11}{l}{\textbf{power mean}  \scriptsize{\lbrack p-values\rbrack}} \\
\midrule
BV \scriptsize{$\lbrack -\infty, 1, \infty\rbrack$}  & 68.7 & 48.0 & 68.8 & 65.8 & 63.7 & 72.2 & 82.5 & 81.3 & 66.9 & 69.5  \\
& \scriptsize{(4.3)} & \scriptsize{(4.7)} & \scriptsize{(1.5)} & \scriptsize{(4.9)} &  \scriptsize{(3.5)} & \scriptsize{(1.4)} & \scriptsize{(3.7)} & \scriptsize{(3.6)} & \scriptsize{(3.9)} & \scriptsize{(11.1)} \\ 
AR \scriptsize{$\lbrack-\infty, 1, \infty\rbrack$} & 71.1 & 44.2 & 67.8 & 68.7 & 68.8 & 75.5 & 84.3 & 84.4 & 73.0 & 73.5 \\
& \scriptsize{(4.5)} & \scriptsize{(8.1)} & \scriptsize{(1.2)} & \scriptsize{(5.0)} &  \scriptsize{(3.8)} & \scriptsize{(2.8)} & \scriptsize{(3.1)} & \scriptsize{(2.5)} & \scriptsize{(4.9)} & \scriptsize{(8.8)} \\
FT \scriptsize{$\lbrack-\infty, 1, \infty\rbrack$} & 69.4 & 43.9 & 64.2 & 69.4 & 67.6 & 73.4 & 85.8 & 81.4 & 73.2 & 65.5 \\
& \scriptsize{(6.2)} & \scriptsize{(9.7)} & \scriptsize{(2.5)} & \scriptsize{(4.4)} &  \scriptsize{(5.8)} & \scriptsize{(3.0)} & \scriptsize{(3.7)} & \scriptsize{(5.1)} & \scriptsize{(5.3)} & \scriptsize{(16.4)} \\
BV $\oplus$ AR $\oplus$ FT \scriptsize{$\lbrack-\infty, 1, \infty\rbrack$} & 73.2 & 50.2 & \textbf{69.3} & \textbf{71.5} & 70.4 & \textbf{76.7} & 86.7 & 84.5 & 75.2 & 74.3 \\
 & \scriptsize{(5.0)} & \scriptsize{(6.8)} & \scriptsize{(1.5)} & \scriptsize{(3.8)} &  \scriptsize{(5.0)} & \scriptsize{(2.4)} & \scriptsize{(4.1)} & \scriptsize{(3.8)} & \scriptsize{(5.9)} & \scriptsize{(12.0)} \\
BV $\oplus$ AR $\oplus$ FT \scriptsize{$\lbrack-\infty, 1, 3, \infty\rbrack$} & \underline{\textbf{73.6}} & \underline{\textbf{52.5}} & 69.1 & 71.1 & \underline{\textbf{70.6}} & \textbf{76.7} & \underline{\textbf{87.5}} & \underline{\textbf{84.9}} & \underline{\textbf{75.5}} & \underline{\textbf{74.8}} \\
 & \scriptsize{(5.0)} & \scriptsize{(5.7)} & \scriptsize{(1.6)} & \scriptsize{(4.3)} &  \scriptsize{(5.2)} & \scriptsize{(2.7)} & \scriptsize{(4.0)} & \scriptsize{(3.3)} & \scriptsize{(5.1)} & \scriptsize{(12.8)} \\
\midrule
\multicolumn{11}{l}{\textbf{Baselines}} \\
\midrule
AR + SIF & 68.1 & 38.4 & 67.7 & \textbf{69.1} & 67.7 & 73.8 & 81.6 & \textbf{81.7} & 70.0 & 63.2 \\
& \scriptsize{(3.5)} & \scriptsize{(3.8)} & \scriptsize{(2.6)} & \scriptsize{(3.0)} &  \scriptsize{(4.2)} & \scriptsize{(1.9)} & \scriptsize{(2.9)} & \scriptsize{(3.1)} & \scriptsize{(6.2)} & \scriptsize{(3.5)} \\
CVM-add & 67.4 & 47.8 & 68.9 & 64.2 & 63.4 & 70.3 & 79.5 & 79.3 & 70.2 & 67.8 \\
& \scriptsize{(5.7)} & \scriptsize{(5.7)} & \scriptsize{(-0.1)} & \scriptsize{(5.9)} &  \scriptsize{(4.5)} & \scriptsize{(5.5)} & \scriptsize{(6.8)} & \scriptsize{(4.5)} & \scriptsize{(6.8)} & \scriptsize{(12.1)} \\
InferSent MT & 71.0 & 49.3 & 69.8 & 67.9 & 69.2 & \textbf{76.3} & \textbf{84.6} & 76.4 & \textbf{73.4} & 72.3  \\
& \scriptsize{(7.4)} & \scriptsize{(8.5)} & \scriptsize{(2.7)} & \scriptsize{(7.3)} &  \scriptsize{(5.1)} & \scriptsize{(4.5)} & \scriptsize{(3.8)} & \scriptsize{(11.7)} & \scriptsize{(6.4)} & \scriptsize{(16.5)} \\
InferSent TD & 71.0 & \textbf{51.1} & \underline{\textbf{72.0}} & 67.9 & 68.9 & 74.7 & 84.3 & 76.8 & 72.7 & 71.0  \\
& \scriptsize{(6.9)} & \scriptsize{(8.3)} & \scriptsize{(1.2)} & \scriptsize{(7.3)} &  \scriptsize{(5.2)} & \scriptsize{(4.2)} & \scriptsize{(4.0)} & \scriptsize{(10.5)} & \scriptsize{(6.1)} & \scriptsize{(15.0)} \\
InferSent MT+TD & \textbf{71.3} & 50.2 & 71.3 & 67.7 & \textbf{69.6} & 76.2 & 84.4 & 77.0 & 72.1 & \textbf{73.2}  \\
& \scriptsize{(7.5)} & \scriptsize{(8.3)} & \scriptsize{(2.2)} & \scriptsize{(8.2)} &  \scriptsize{(5.4)} & \scriptsize{(5.1)} & \scriptsize{(4.3)} & \scriptsize{(11.1)} & \scriptsize{(7.1)} & \scriptsize{(15.9)} \\
\bottomrule
\end{tabular}
\caption{Cross-lingual results averaged over en$\rightarrow$de and en$\rightarrow$fr. Numbers in parentheses are the in-language results minus the given cross-language value.
}
\label{table:results:x-ling}
\end{table*}

For ease of consideration, we report average results 
over en$\rightarrow$de and en$\rightarrow$fr in Table \ref{table:results:x-ling}. 
Per-language scores can be found in the appendix.

As in the monolingual case, we observe  substantial improvements when concatenating different types of word embeddings of $\sim$2pp on average.
However, when adding FT embeddings to the already strong concatenation BV$\oplus$AR, %however, 
the 
performance only slightly improves on average.

Conversely, using different power means is more  
effective, considerably improving 
performance values compared to arithmetic mean word embeddings. 
On average, concatenation of word embedding types plus different power means beats the best individual word embeddings by 4.4pp cross-lingually, 
from 69.2\% for AR to 73.6\%. 

We not only beat all our InferSent adaptations by more than 2pp on average cross-lingually, 
our concatenated power mean embeddings also outperform the more complex InferSent adaptations %also 
in 8 out of 9 individual transfer tasks. %transfer tasks. 

Further, we 
perform on par with InferSent already with dimensionality $d=900$, either using the concatenation of our three cross-lingual word embeddings or using AR with three power means ($p=1,\pm\infty$). In contrast, CVD-add and AR$+$SIF stay below InferSent, and, as in the monolingual case, even underperform relative to the best individual cross-lingual average word embedding baseline (AR), indicating that they are not suitable as universal feature representations. 

\section{Analysis}
\label{sec:analysis}

\paragraph{Machine translations}

To test the validity of our evaluations that are 
based on machine translations, we %also 
compared 
performances
when evaluating on machine (\textsf{MT}) and human translations (\textsf{HT}) of our two parallel AM and AC datasets.

We re-evaluated the same 14 methods as in Table \ref{table:results:x-ling} using \textsf{MT} target data. We find a Spearman correlation of $\rho=$ 96.5\% and a Pearson correlation of $\tau=$ 98.4\% between \textsf{HT} and \textsf{MT} for AM. 
For AC we find a $\rho$ value of 83.7\% and a $\tau$ value of 89.9\%. While the latter correlations are lower, we note that the AC scores are rather close in the direction en$\rightarrow$de, so small changes (which may also be due to chance, given the dataset's small size) can lead to rank differences.
Overall, 
this indicates
that our \textsf{MT} experiments yield reliable rankings 
and that they strongly correlate to performance values measured on \textsf{HT}. Indeed, introspecting the 
machine translations, we observed that these were of very high perceived quality. 

\paragraph{Different power means}
\label{sec:discussion:p-values}

We performed additional cross-lingual experiments based on the concatenation of BV $\oplus$ AR $\oplus$ FT with additional $p$-means. In particular, we test (i) if some power means are more effective than others, and (ii) if using more power means, and thus increasing the dimensionality of the embeddings, further increases performances.

We chose several intuitive values for power mean in addition to the ones already tried out, namely $p=-1$ (harmonic mean), $p=0.5$, $p=2$ (quadratic mean), and $p=3$ (cubic mean). 
Table \ref{table:results:p-values} reports the average performances over all tasks. We notice that
$p=3$ is the most effective power mean here and $p=-1$ is (by far) least effective. 
We discuss below why $p=-1$ hurts performances in this case.
For all cases with $p > 0$, 
%the addition of multiple means tends
additional means tend
 to further improve the results, but with decreasing marginal returns.
%These results also mean
This also means that improvements are not merely due to additional dimensions 
%(indeed, adding $p=-1$ hurts performances) 
but due to addition of complementary
%, useful 
information. 
\begin{table}
\centering
\footnotesize
\begin{tabular}{lcc}
\toprule
\textbf{power mean-values} & \textbf{$\Sigma$ X-Ling} & \textbf{$\Sigma$ In-Language} \\
\midrule
$p=1,\pm\infty$ & 73.2 & 78.2 \\
\midrule
$p=1,\pm\infty,-1$ %\scriptsize{(harmonic mean)} 
& 59.9 & 61.6 \\
$p=1,\pm\infty,0.5$ & 73.0 & 78.6  \\
$p=1,\pm\infty,2$ %\scriptsize{(quadratic mean)} 
& 73.4 & 78.5 \\
$p=1,\pm\infty,3$ %\scriptsize{(cubic mean)} 
& 73.6 & 78.6 \\
\midrule
$p=1,\pm\infty,2,3$ &  \textbf{73.7} & 78.7 \\
$p=1,\pm\infty,0.5,2,3$ & 73.6 & \textbf{78.9} \\
\bottomrule
\end{tabular}
\caption{Average scores (en$\rightarrow$de and en$\rightarrow$fr) for additional power means (based on BV $\oplus$ AR $\oplus$ FT). 
}
\label{table:results:p-values}
\end{table}

\section{Discussion}
\label{sec:discussion}

\paragraph{Why is it useful to concatenate power means?}
The average of word embeddings 
discards a lot of information because different sentences can be represented by similar averages. The concatenation of different power means 
yields a more precise  
summary 
because it
reduces uncertainty about the semantic variation within a sentence.
For example, knowing the min and the max guarantees that  
embedding dimensions are all within certain ranges.

\paragraph{Which power means promise to be beneficial?}
Large $|p|$ quickly converge to min ($p = -\infty$) and max ($p = \infty$). Hence, besides min and max, 
further good power mean-values are typically small numbers, e.g., $|p|<10$. If they are integral, then odd numbers are preferable over even ones because even power means lose sign information. Further, positive power means are preferable over negative ones (see results in Table \ref{table:results:p-values}) because negative power means are in a fundamental sense discontinuous when input numbers are negative: they have multiple poles (power mean value tends toward $\pm\infty$) and different signs around the poles, depending on the direction from which one approaches the pole.   

\paragraph{Cross-lingual performance decrease}
For all models and tasks, we observed decreased performances in the cross-lingual evaluation compared to the in-language evaluation. Most importantly, we observe a substantial difference between the performance decreases of our best model (5pp) and our best cross-lingual InferSent adaptation (7.5pp). 
Two reasons may explain these observations. 

First, InferSent is a 
complex approach based on a bidirectional LSTM. 
In the same vein as 
\newcite{Wieting16}, we hypothesize that embeddings learned from complex approaches transfer less well across domains compared to embeddings derived from simpler methods such as power mean embeddings. 
In our case, we transfer across languages, which is a pronounced form of domain shift.

Second, InferSent typically requires large amounts of high-quality training data. In the cross-lingual case we rely on translated sentences for training.
Even though
we found these translation to be of high quality, 
they can still introduce noise because some aspects of meaning in languages can only be approximately captured by translations. 
This effect could increase with 
more
distance between languages. 
In particular, we observe a higher cross-language drop for the language transfer en$\rightarrow$fr than for 
en$\rightarrow$de. Furthermore, this difference is less pronounced for our 
power mean embeddings than it is for InferSent, potentially supporting this assumption (see the appendix for individual cross-language results).

\paragraph{Applications for cross-lingual embeddings}
A fruitful application scenario of cross-lingual sentence embeddings are cases in which we do not have access to labeled target language training data. 
Even though we could, in theory, machine translate sentences from the target language into English and apply a monolingual classifier, state-of-the-art MT systems like Google Translate currently only cover 
a small fraction of the world's $\sim$7000 languages. 

Using cross-lingual sentence embeddings, however, we can train a classifier on English and then directly apply it to low-resource target language sentences.
This so-called direct transfer approach 
%\cite{McDonald:2011,Zhang:2016} 
\cite{Zhang:2016} 
on sentence-level can be beneficial when labeled data is scarce, because sentence-level approaches typically outperform task-specific sentence representations induced from word-level models in this case \cite{Subramanian2018}.

\section{Conclusion}\label{sec:conclusion}

We proposed concatenated power mean word embeddings, a conceptually and computationally simple 
method for inducing sentence embeddings 
using two ingredients: (i) the concatenation of
diverse word embeddings, which injects complementary information in the resulting representations; (ii) the use of power means to perform different types of summarizations over word embedding dimensions.

Our proposed method narrows the monolingual gap to 
state-of-the-art supervised methods
while substantially outperforming cross-lingual adaptations of InferSent in the cross-lingual scenario.

We believe that our generalizations can be widely extended and that we have merely laid the conceptual ground-work: 
automatically learning power mean-values is likely to result in further improved sentence embeddings as we have observed that some power mean-values are more suitable than others, and using different power mean-values for different embedding dimensions could introduce more diversity.

Finally, we believe that even in monolingual scenarios, 
future work should consider (concatenated) power mean embeddings 
as a challenging and truly hard-to-beat baseline
across a wide array of transfer tasks.

\section*{Acknowledgments}
This work has been supported by the German Research Foundation as part of the
QA-EduInf project (grant GU 798/18-1 and grant RI 803/12-1), by the German
Federal Ministry of Education and Research (BMBF) under
the promotional reference 01UG1816B (CEDIFOR), 
and 
by the German
Research Foundation (DFG) as part of the Research
Training Group ``Adaptive Preparation of Information
from Heterogeneous Sources'' (AIPHES)
under grant No.\ GRK 1994/1.

\bibliography{paper}
\bibliographystyle{acl_natbib}

\appendix
\clearpage

%\newpage

\section{Supplemental Material}

\subsection{Details on our Projection Method}
\label{appendix:mapping}

Here we describe the necessary conceptual and technical details to reproduce the results of our non-linear projection method that we use to map word embeddings of two languages into a shared embedding space (cf. \S\ref{sec:xling-exp}).

\paragraph{Formalization}
We learn a projection of two embedding spaces $\mathbb{E}^{l}$ and $\mathbb{E}^{k}$ with dimensionality $e$ and $f$, respectively, into a shared space of dimensionality $d$ using two non-linear transformations:
\begin{align*}
	f_{l} (\mathbf{x}_{l}) &= \tanh \left( \mathbf{W}_{l} \mathbf{x}_l + \mathbf{b}_{l} \right) \\
    f_{k} (\mathbf{x}_{k}) &= \tanh \left( \mathbf{W}_{k} \mathbf{x}_k + \mathbf{b}_{k} \right)
\end{align*}
where $\mathbf{x}_{l} \in \mathbb{R}^{e}$ , $ \mathbf{x}_{k} \in \mathbb{R}^{f}$ are original input embeddings and $\mathbf{W}_{l}\in\mathbb{R}^{d\times e}, \mathbf{W}_{k} \in \mathbb{R}^{d \times f}$, $\mathbf{b}_{l}\in\mathbb{R}^{d}, \mathbf{b}_{k} \in \mathbb{R}^{d}$ are parameters to be learned. Here $\mathbf{x}_l$ and $\mathbf{x}_k$ 
are monolingual representations.

For each sentence $s$ and its translation $t$ we randomly sample one unrelated sentence $u$ from
our data and obtain sentence representations $\mathbf{r}_s = f_{l}(\mathbf{x}_s)$, $\mathbf{r}_t = f_{k}(\mathbf{x}_t)$, and $\mathbf{r}_u = f_{k}(\mathbf{x}_u)$. 
We then optimize the following max-margin hinge loss:
\begin{equation*}
	\mathcal{L} = \max\left(0,~ m-\mathsf{sim}(\mathbf{r}_s, \mathbf{r}_t) + \mathsf{sim}(\mathbf{r}_s, \mathbf{r}_u)\right)
\end{equation*}
where $\mathsf{sim}$ is cosine similarity and $m$ is the margin parameter. This objective moves embeddings of translations closer to and embeddings of random cross-lingual sentences further away from each other.

\paragraph{Training}
We use minibatched SGD with the Adam optimizer \cite{Kingma2015} for training. We train on $>$130K  bilingually aligned sentence pairs from the TED corpus \cite{Hermann2014}, which consists of translated transcripts from TED talks. Each sentence $s$ is represented by its average (monolingual) word embedding, i.e., $H_1$. 

We set the margin parameter to $m = 0.5$ as we have observed that higher values lead to a faster convergence. 
We furthermore randomly set 50\% of the input embedding dimensions to zero during training (dropout).

Training of one epoch usually takes less than a minute in our TensorFlow implementation (on CPU), and convergence is usually achieved after less than 100 epochs.

\paragraph{Application}

\begin{table}
\centering
\footnotesize
\begin{tabular}{lcc}
\toprule
\textbf{Model} & $\sum$ \textbf{X-Ling} & $\sum$ \textbf{In-Language} \\
\midrule
FT (monolingual) & - & 80.8 \\
FT (CCA$^\ddagger$) & 71.1 & 79.3 \\
FT (our projection) & 74.6 & 79.7 \\
\midrule
BV (orig) & 70.9 & 75.8 \\
BV (our projection) & 71.0 & 74.6 \\
\midrule
AR (orig) & 61.8 & 79.3 \\
AR (our projection) & 74.5 & 77.9 \\
\bottomrule
\end{tabular}
\caption{The performance of our average word embeddings with our projection method in comparison to other approaches. $^\ddagger$We trained CCA on word-alignments extracted from TED transcripts using fast\_align (i.e., CCA uses the same data source as our method).}
\label{table:appendix:mapping-results}
\end{table}

Even though we learn our non-linear projection on the sentence level, we later apply it on the word level, i.e., we map individual word embeddings from each of  two languages via $f_{\psi}(\mathbf{x}_\psi)$ where $\psi=l,k$. This is valid because average word embeddings live in the same space as individual word embeddings. The reason for doing so is that otherwise we would have to learn individual transformations for each of our power means, not only the average ($=$ $H_1$),
which would be too costly particularly when incorporating many different $p$-values. 
Working on the word-level, in general, also allows us to resort to word-level projection techniques using, e.g., word-alignments rather than sentence alignments. 

However, in preliminary experiments, we found that our suggested approach produces considerably better cross-lingual word embeddings in our setup. Results are shown in Table \ref{table:appendix:mapping-results}, where we report the performance of average word embeddings for cross-lingual en$\rightarrow$de task transfer (averaged over MR, CR, SUBJ, MPQA, SST, TREC). Compared to the word-level projection method CCA we obtain substantially better cross-lingual sentence embeddings, and even stronger improvements when re-mapping 
AR embeddings, even though these are already bilingual.

\subsection{Individual Language-Transfer Results}
\label{appendix:fine-grained-results}

We report results for the individual language transfer across en$\rightarrow$de and en$\rightarrow$fr in Table \ref{table:appendix:full}.

\begin{sidewaystable*}
%\begin{table*}
\centering
\hspace{-2.5cm}
\footnotesize
\begin{tabular}{l|c|cc||cc|cc|cc|cc|cc|cc|cc|cc|cc}
\toprule
\textbf{Model} & $\Sigma$ &  $\Sigma$ \textbf{de} &  $\Sigma$ \textbf{fr} & \multicolumn{2}{c|}{\textbf{AM}} & \multicolumn{2}{c|}{\textbf{AC}} & \multicolumn{2}{c|}{\textbf{CLS}} & \multicolumn{2}{c|}{\textbf{MR}} & \multicolumn{2}{c|}{\textbf{CR}} & \multicolumn{2}{c|}{\textbf{SUBJ}} & \multicolumn{2}{c|}{\textbf{MPQA}} & \multicolumn{2}{c|}{\textbf{SST}} & \multicolumn{2}{c}{\textbf{TREC}} \\ 
Transfer Language & & & & de & fr & de & fr & de & fr & de & fr & de & fr & de & fr & de & fr & de & fr & de & fr \\
\midrule 
\multicolumn{11}{l}{\textbf{Arithmetic mean}} \\
\midrule
BIVCD (BV) & 67.3 & 65.5 & 68.1 & 39.2 & 41.9 & 68.9 & 66.4 & 65.0 & 67.7 & 62.2 & 66.5 & 70.6 & 72.9 & 79.8 & 82.4 & 79.8 & 83.3 & 61.2 & 70.2 & 72.2 & 61.8 \\
 & \xldrop{3.7} & \xldrop{3.9} & \xldrop{3.6} & \xldrop{7.0} & \xldrop{4.8} & \xldrop{1.5} & \xldrop{4.7} & \xldrop{4.3} & \xldrop{3.7} & \xldrop{2.3} & \xldrop{1.5} & \xldrop{1.1} & \xldrop{0.1} &\xldrop{3.8} & \xldrop{3.2} &\xldrop{3.9} & \xldrop{2.4} &\xldrop{7.2} & \xldrop{0.3} &\xldrop{3.6} & \xldrop{11.8}\\
Attract-Repel (AR) & 69.2 & 69.4 & 68.9 & 39.0 & 38.2 & \textbf{71.1} & 66.5 & 67.4 & 70.4 & 66.6 & 69.7 & 73.7 & 74.0 & 81.8 & 83.8 & 84.0 & 84.8 & 71.3 & 73.6 & 69.6 & 59.4  \\
 & \xldrop{3.6} & \xldrop{3.4} & \xldrop{3.8} & \xldrop{4.7} & \xldrop{4.7} & \xldrop{0.4} & \xldrop{1.3} & \xldrop{5.8} & \xldrop{2.8} & \xldrop{4.7} & \xldrop{2.2} & \xldrop{1.9} & \xldrop{2.4} & \xldrop{3.2} & \xldrop{2.7} & \xldrop{2.1} & \xldrop{1.5} & \xldrop{4.4} & \xldrop{2.4} & \xldrop{3.8} & \xldrop{14.6} \\
FastText (FT) & 68.4 & 68.7 & 68.0 & 36.9 & 40.0 & 63.8 & 62.9 & 70.1 & 70.0 & 68.3 & 69.9 & 73.9 & 72.3 & 86.3 & 84.0 & 81.7 & 81.3 & 69.5 & 69.2 & 67.8 & 62.4 \\
 & \xldrop{5.6} & \xldrop{5.5} & \xldrop{5.7} & \xldrop{10.5} & \xldrop{6.6} & \xldrop{4.0} & \xldrop{1.8} & \xldrop{4.2} & \xldrop{4.0} & \xldrop{5.2} & \xldrop{3.0} & \xldrop{2.0} & \xldrop{3.1} & \xldrop{2.4} & \xldrop{4.8} & \xldrop{4.5} & \xldrop{4.5} & \xldrop{8.8} & \xldrop{8.5} & \xldrop{8.2} & \xldrop{15.2} \\
BV $\oplus$ AR & 71.1 & 70.9 & \textbf{71.3} & \textbf{40.8} & \textbf{42.3} & 70.0 & \textbf{67.6} & 69.4 & 70.9 & 67.2 & 70.5 & 75.4 & 76.5 & 83.3 & 84.8 & 84.1 & \textbf{85.4} & \textbf{72.5} & \textbf{75.6} & \textbf{75.8} & \textbf{68.0} \\
 & \xldrop{4.2} & \xldrop{4.2} & \xldrop{4.1} & \xldrop{9.7} & \xldrop{7.6} & \xldrop{2.2} & \xldrop{3.9} & \xldrop{4.6} & \xldrop{2.9} & \xldrop{4.7} & \xldrop{2.1} & \xldrop{1.1} & \xldrop{0.4} & \xldrop{3.8} & \xldrop{3.1} & \xldrop{3.3} & \xldrop{2.5} & \xldrop{5.3} & \xldrop{3.0} & \xldrop{3.4} & \xldrop{11.6} \\
BV $\oplus$ AR $\oplus$ FT & \textbf{71.2} & \textbf{71.9} & 70.6 & 39.2 & 40.7 & 71.0 & 64.5 & \underline{\textbf{71.2}} & \underline{\textbf{72.1}} & \textbf{69.8} & \underline{\textbf{70.8}} & \textbf{76.6} & \underline{\textbf{76.9}} & \textbf{86.8} & \textbf{85.7} & \underline{\textbf{84.6}} & 84.9 & 71.8 & 74.7 & \textbf{75.8} & 65.2 \\
 & \xldrop{5.9} & \xldrop{5.4} & \xldrop{6.4} & \xldrop{12.9} & \xldrop{11.0} & \xldrop{1.5} & \xldrop{5.0} & \xldrop{3.8} & \xldrop{3.4} & \xldrop{6.1} & \xldrop{4.0} & \xldrop{1.6} & \xldrop{1.2} & \xldrop{3.2} & \xldrop{4.4} & \xldrop{3.3} & \xldrop{3.3} & \xldrop{8.5} & \xldrop{5.1} & \xldrop{7.8} & \xldrop{20.0} \\
\midrule
\multicolumn{11}{l}{\textbf{$p$-mean}  \scriptsize{\lbrack p-values\rbrack}} \\
\midrule
BV \scriptsize{$\lbrack -\infty, 1, \infty\rbrack$}  & 68.7 & 68.0 & 69.5 & 48.0 & 47.9 & 70.7 & \textbf{66.8} & 64.4 & 67.3 & 60.5 & 66.8 & 71.1 & 73.3 & 81.1 & 83.9 & 79.9 & 82.7 & 64.4 & 69.4 & 72.0 & 67.0 \\
 & \xldrop{4.3} & \xldrop{4.1} & \xldrop{4.4} & \xldrop{4.1} & \xldrop{5.4} & \xldrop{0.8} & \xldrop{2.3} & \xldrop{5.5} & \xldrop{4.2} & \xldrop{4.2} & \xldrop{2.7} & \xldrop{2.2} & \xldrop{0.6} & \xldrop{4.4} & \xldrop{2.9} & \xldrop{3.9} & \xldrop{3.3} & \xldrop{3.1} & \xldrop{4.7} & \xldrop{8.4} & \xldrop{13.8} \\
AR \scriptsize{$\lbrack-\infty, 1, \infty\rbrack$} & 71.1 & 71.3 & 71.0 & 45.7 & 42.7 & 70.5 & 65.1 & 67.1 & 70.3 & 67.4 & 70.2 & 75.3 & 75.7 & 83.7 & 84.9 & 84.0 & 84.8 & 71.2 & 74.9 & 76.6 & 70.4 \\
 & \xldrop{4.5} & \xldrop{4.4} & \xldrop{4.5} & \xldrop{6.7} & \xldrop{9.1} & \xlgain{0.1} & \xldrop{2.6} & \xldrop{6.6} & \xldrop{3.5} & \xldrop{5.2} & \xldrop{2.3} & \xldrop{3.5} & \xldrop{2.1} & \xldrop{3.2} & \xldrop{3.1} & \xldrop{2.7} & \xldrop{2.3} & \xldrop{6.6} & \xldrop{3.2} & \xldrop{5.6} & \xldrop{12.0} \\
FT \scriptsize{$\lbrack-\infty, 1, \infty\rbrack$} & 69.4 & 70.2 & 68.5 & 42.7 & 45.1 & 67.1 & 61.3 & 69.6 & 69.2 & 68.3 & 67.0 & 73.4 & 73.4 & 86.7 & 84.9 & 81.6 & 81.2 & \underline{\textbf{74.4}} & 72.0 & 68.2 & 62.8 \\
 & \xldrop{6.2} & \xldrop{5.4} & \xldrop{7.0} & \xldrop{11.1} & \xldrop{8.2} & \xldrop{1.3} & \xldrop{3.7} & \xldrop{4.2} & \xldrop{4.6} & \xldrop{4.9} & \xldrop{6.6} & \xldrop{3.0} & \xldrop{3.0} & \xldrop{2.7} & \xldrop{4.7} & \xldrop{5.2} & \xldrop{5.0} & \xldrop{4.0} & \xldrop{6.7} & \xldrop{12.0} & \xldrop{20.8} \\
BV $\oplus$ AR $\oplus$ FT \scriptsize{$\lbrack-\infty, 1, \infty\rbrack$} & 73.2 & 73.7 & 72.7 & 50.8 & 49.6 & \textbf{72.0} & 66.6 & \textbf{70.9} & \textbf{72.0} & \underline{\textbf{70.6}} & 70.2 & \underline{\textbf{77.3}} & 76.1 & 86.6 & 86.8 & 84.1 & 84.9 & 73.4 & \underline{\textbf{77.0}} & 77.6 & 71.0 \\
 & \xldrop{5.0} & \xldrop{4.6} & \xldrop{5.5} & \xldrop{5.6} & \xldrop{7.7} & \xldrop{0.3} & \xldrop{2.8} & \xldrop{4.1} & \xldrop{3.4} & \xldrop{4.7} & \xldrop{5.2} & \xldrop{2.3} & \xldrop{2.5} & \xldrop{4.2} & \xldrop{4.0} & \xldrop{3.9} & \xldrop{3.6} & \xldrop{8.1} & \xldrop{3.7} & \xldrop{7.8} & \xldrop{16.8} \\
BV $\oplus$ AR $\oplus$ FT \scriptsize{$\lbrack-\infty, 1, 3, \infty\rbrack$} & \underline{\textbf{73.6}} & \underline{\textbf{74.0}} & \underline{\textbf{73.3}} & \textbf{51.4} & \underline{\textbf{53.6}} & \textbf{72.0} & 66.3 & 70.8 & 71.5 & 70.5 & \textbf{70.7} & 77.1 & \textbf{76.2} & \underline{\textbf{87.7}} & \underline{\textbf{87.3}} & \textbf{84.2} & \underline{\textbf{85.6}} & 74.1 & 76.9 & \textbf{78.4} & \underline{\textbf{71.2}} \\
 & \xldrop{5.0} & \xldrop{4.5} & \xldrop{5.3} & \xldrop{6.6} & \xldrop{4.8} & \xldrop{0.5} & \xldrop{2.7} & \xldrop{4.3} & \xldrop{3.8} & \xldrop{5.5} & \xldrop{4.9} & \xldrop{2.7} & \xldrop{2.6} & \xldrop{3.6} & \xldrop{4.3} & \xldrop{3.8} & \xldrop{2.8} & \xldrop{6.6} & \xldrop{3.6} & \xldrop{7.2} & \xldrop{17.8} \\
\midrule
\multicolumn{11}{l}{\textbf{Baselines}} \\
\midrule
AR + SIF & 68.1 & 67.5 & 68.7 & 40.2 & 36.5 & 70.3 & 65.1 & \textbf{67.7} & \textbf{70.4} & 66.2 & 69.2 & 73.7 & 73.8 & 80.0 & 83.2 & \textbf{82.9} & 80.5 & 68.0 & 71.9 & 58.4 & 68.0 \\
& \xldrop{3.5} & \xldrop{4.3} & \xldrop{2.7} & \xldrop{2.3} & \xldrop{5.2} & \xldrop{1.8} & \xldrop{3.4} & \xldrop{4.5} & \xldrop{1.5} & \xldrop{5.5} & \xldrop{3.0} & \xldrop{2.3} & \xldrop{1.6} & \xldrop{3.8} & \xldrop{2.0} &  \xldrop{2.0} & \xldrop{4.1} & \xldrop{8.8} & \xldrop{3.7} & \xldrop{7.4} & \xlgain{0.4} \\
CVM-Add & 67.4 & 65.3 & 69.4 & 45.6 & \textbf{49.9} & 69.6 & 68.1 & 62.3 & 66.0 & 60.7 & 66.1 & 68.6 & 72.0 & 76.6 & 82.3 & 76.1 & \textbf{82.5} & 62.7 & 67.7 & 65.2 & \textbf{70.4} \\
& \xldrop{5.7} & \xldrop{6.6} & \xldrop{4.8} & \xldrop{7.7} & \xldrop{3.7} & \xlgain{0.7} & \xlgain{0.1} & \xldrop{6.3} & \xldrop{5.5} & \xldrop{4.9} & \xldrop{4.0} & \xldrop{6.5} & \xldrop{4.5} & \xldrop{8.5} & \xldrop{5.0} &  \xldrop{5.7} & \xldrop{3.3} & \xldrop{7.7} & \xldrop{5.9} & \xldrop{13.2} & \xldrop{11.0} \\
InferSent MT & 71.0 & 71.8 & \textbf{70.2} & 50.3 & 48.3 & 70.9 & 68.7 & 67.0 & 68.9 & 69.3 & 69.2 & \textbf{76.7} & 75.8 & 84.4 & \textbf{84.9} & 77.9 & 74.9 & \textbf{72.5} & \textbf{74.3} & 77.4 & 67.2 \\
 & \xldrop{7.4} & \xldrop{6.7} & \xldrop{8.2} & \xldrop{8.2} & \xldrop{9.5} & \xldrop{1.2} & \xldrop{4.2} & \xldrop{8.5} & \xldrop{6.1} & \xldrop{5.2} & \xldrop{4.9} & \xldrop{3.5} & \xldrop{5.5} & \xldrop{3.7} & \xldrop{3.8} & \xldrop{10.2} & \xldrop{13.3} & \xldrop{7.4} & \xldrop{5.5} & \xldrop{12.2} & \xldrop{20.8} \\
InferSent TD & 71.0 & 72.1 & 70.0 & \underline{\textbf{52.7}} & 49.5 & 73.4 & \underline{\textbf{70.6}} & 66.8 & 69.1 & 68.6 & 69.2 & 74.9 & 74.4 & 84.3 & 84.2 & 77.4 & 76.3 & \textbf{72.5} & 72.8 & 78.2 & 63.8 \\
 & \xldrop{6.9} & \xldrop{5.9} & \xldrop{7.9} & \xldrop{7.0} & \xldrop{10.0} & \xldrop{1.7} & \xldrop{0.7} & \xldrop{8.9} & \xldrop{5.7} & \xldrop{4.7} & \xldrop{5.8} & \xldrop{3.8} & \xldrop{4.6} & \xldrop{3.7} & \xldrop{4.3} & \xldrop{9.7} & \xldrop{11.3} & \xldrop{5.9} & \xldrop{6.3} & \xldrop{7.4} & \xldrop{22.6} \\
InferSent MT+TD & \textbf{71.3} & \textbf{72.4} & \textbf{70.2} & 52.0 & 48.4 & \underline{\textbf{73.5}} & 69.2 & 66.6 & 68.8 & \textbf{69.5} & \textbf{69.7} & 76.4 & \textbf{76.0} & \textbf{84.7} & 84.2 & 78.3 & 75.7 & 72.1 & 72.2 & \underline{\textbf{78.8}} & 67.6 \\
 & \xldrop{7.5} & \xldrop{6.6} & \xldrop{8.4} & \xldrop{7.0} & \xldrop{9.6} & \xldrop{1.4} & \xldrop{3.0} & \xldrop{9.6} & \xldrop{6.8} & \xldrop{5.2} & \xldrop{5.6} & \xldrop{4.8} & \xldrop{5.5} & \xldrop{4.2} & \xldrop{4.4} & \xldrop{9.7} & \xldrop{12.4} & \xldrop{7.1} & \xldrop{7.0} & \xldrop{10.8} & \xldrop{21.0} \\
\bottomrule
\end{tabular}
\caption{Individual cross-lingual results for the language transfer en$\rightarrow$de and en$\rightarrow$fr. Numbers in parentheses are the in-language results minus the given cross-language value.. $\oplus$ denotes the concatenation of different embeddings (or $p$-means), brackets show the different $p$-means of the model. 
}
\label{table:appendix:full}
\end{sidewaystable*}
%\end{table*}

\end{document}